\documentclass[runningheads]{llncs}

\usepackage{multirow}
\usepackage{graphicx}

\usepackage{float}
\usepackage{placeins}

\usepackage{tabularx}
\usepackage{threeparttable}
\usepackage{ragged2e}
\newcolumntype{Y}{>{\RaggedRight\arraybackslash}X}
\usepackage{eccvabbrv}
\usepackage{graphicx}
\usepackage{booktabs}
\usepackage{threeparttable}
\usepackage{adjustbox}
\usepackage{pifont}
\newcommand{\cmark}{\ding{51}}
\newcommand{\xmark}{\ding{55}}

\newcommand{\best}[1]{\textbf{#1}}
\newcommand{\second}[1]{\underline{#1}}

\usepackage[ruled,vlined]{algorithm2e}
\usepackage{placeins}



\usepackage[accsupp]{axessibility}  

\usepackage{hyperref}

\usepackage{orcidlink}

\begin{document}


\title{Single Image Super-Resolution via Bivariate \`A Trous Wavelet Diffusion} 

\titlerunning{BATDiff}

\author{Maryam Heidari\inst{1}\orcidlink{0000-1111-2222-3333} \and
Nantheera Anantrasirichai\inst{1}\orcidlink{1111-2222-3333-4444} \and
Alin Achim\inst{1}\orcidlink{2222--3333-4444-5555}}

\authorrunning{M. Heidari et al.}

\institute{Visual Information Laboratory, University of Bristol, Bristol, UK\\
\email{il24997@bristol.ac.uk, n.anantrasirichai@bristol.ac.uk, alin.achim@bristol.ac.uk}
}
\maketitle

\begin{abstract}

The effectiveness of super-resolution (SR) models hinges on their ability to recover high-frequency structure without introducing artifacts. Diffusion-based approaches have recently advanced the state of the art in SR. However, most diffusion-based SR pipelines operate purely in the spatial domain, which may yield high-frequency details that are not well supported by the underlying low-resolution evidence. On the other hand, unlike supervised SR models that may inject dataset-specific textures, single-image SR relies primarily on internal image statistics and can therefore be less prone to dataset-driven hallucinations; nevertheless, ambiguity in the LR observation can still lead to inconsistent high-frequency details. To tackle this problem, we introduce BATDiff, an unsupervised Bivariate \`A trous Wavelet Diffusion model designed to provide structured cross-scale guidance during the generative process. BATDiff employs an \`a Trous wavelet transform that constructs an undecimated multiscale representation in which high-frequency components are progressively revealed while the full spatial resolution is preserved. As the core inference mechanism, BATDiff includes a bivariate cross-scale module that models parent–child dependencies between adjacent scales. It improves high-frequency coherence and reduces mismatch artifacts in diffusion-based SR. Experiments on standard benchmarks demonstrate that BATDiff produces sharper and more structurally consistent reconstructions than existing diffusion and non-diffusion baselines, achieving improvements in fidelity and perceptual quality.
  
  \keywords{Super-resolution \and diffusion models \and wavelet
transforms}
\end{abstract}

\section{Introduction}
\label{sec:intro}
Single-image super-resolution (SISR) addresses the inverse problem of recovering a high-resolution (HR) image from a single low-resolution (LR) observation, where substantial high-frequency information is irreversibly lost during image acquisition~\cite{Zhang20}. Despite the remarkable progress achieved by deep learning-based approaches, the faithful reconstruction of fine-scale structures remains a persistent challenge, particularly in real-world scenarios characterized by complex degradations such as unknown blur kernels, noise contamination, and non-ideal downsampling processes~\cite{BSRGAN,Zhao2022,Lepcha}. Existing methods often struggle to balance distortion minimization with perceptual fidelity, frequently resulting in either oversmoothed outputs or visually sharp yet structurally inconsistent artifacts~\cite{BlauMichaeli2022}. Consequently, developing SR frameworks capable of restoring high-frequency details while preserving global coherence continues to be a central objective in the field.

Recent advances in generative modelling have transformed SR from deterministic regression into a probabilistic reconstruction problem~\cite{zhai2023}. Generative approaches learn rich priors over natural image statistics and use them to model the conditional distribution of HR images given LR observations. This formulation supports the recovery of fine-scale structures that remain ill-constrained under low-resolution inputs.
 Among these approaches, diffusion probabilistic models have gained increasing attention due to their stable training dynamics and strong performance in image generation and restoration tasks~\cite{DhariwalNichol2021,Song2021,Karras2022}. Diffusion-based methods cast SR within a conditional denoising framework~\cite{Li2025DiffusionSurvey}, where successive refinement steps progressively approximate the target distribution and facilitate structured high-frequency reconstruction.

Despite their strong empirical performance, prevailing diffusion-based SISR formulations typically perform reconstruction at a single effective resolution scale, whether defined in pixel space~\cite{Cui} or a latent/feature space~\cite{StableSR}. From a multiscale perspective, an image can be represented through wavelet decompositions as a collection of subband coefficients at progressively finer frequency levels defined on a common spatial grid~\cite{DiWa}. Within this representation, classical statistical analyses have demonstrated that high-frequency coefficients are conditionally dependent on their corresponding parent coefficients at the adjacent coarser level~\cite{Portilla,Alin,Bivariate,Sendur}. These parent–child relationships capture intrinsic cross-scale structure in natural images. However, such cross-scale statistical dependencies are not explicitly modeled in most existing diffusion-based SISR formulations. Consequently, high-frequency details may be generated without structural alignment to the underlying coarse information inferred from the LR observation. This limitation may become particularly evident in challenging reconstruction scenarios, where stable texture generation requires consistent interaction between low- and high-frequency components~\cite{Lin}.

In this paper, we propose BATDiff, an unsupervised (no external paired LR–HR supervision) diffusion-based framework for SISR that incorporates \`a trous wavelet decomposition to construct a shift-invariant multiscale representation~\cite{Atrous}. Owing to its undecimated formulation, the \`a trous transform preserves full spatial resolution at each decomposition level while maintaining precise spatial alignment between low- and high-frequency subbands. Building upon this aligned multiscale structure, BATDiff introduces a bivariate cross-scale reconstruction strategy within the reverse diffusion process. At each reverse step, the reconstruction at a given frequency band is conditioned not only on its current noisy estimate but also on the time-aligned estimate from the adjacent coarser band. This explicit parent–child conditioning embeds structured cross-scale statistical dependencies directly into diffusion inference and promotes consistency between coarse structural components and generated fine details.

Our contributions can be summarized as follows:

\begin{itemize}
\item We introduce a bivariate cross-scale conditioning mechanism that models multi-scale statistical dependencies within reverse diffusion inference.
\item We employ an \`a trous wavelet decomposition to construct a spatially aligned multiscale representation that enables stable cross-scale conditioning during reconstruction.
\item We develop an unsupervised SISR framework based on internal learning without paired LR–HR supervision while ensuring consistency with observed LR input.
\end{itemize}

Extensive experiments on standard SR benchmarks demonstrate that BATDiff achieves strong reconstruction fidelity at $\times4$, reaching 28.53 dB PSNR and 0.8502 SSIM on the challenging Urban100 dataset~\cite{Urban100}, while recovering finer details and reducing over-smoothing artifacts.

The remainder of the paper is organized as follows. Section~2 presents background and related work. Section~3 and~4 introduce the necessary preliminaries and the proposed BATDiff framework, respectively. Experimental evaluations and ablation analyses are presented in Section~5, followed by conclusions in Section~6.

\section{Related Work}

The trajectory of SR research mirrors broader developments in image restoration, transitioning from supervised regression models to probabilistic generative frameworks. The following discussion situates the proposed approach within this landscape.

\subsection{Learning-Based Super-Resolution}

Early deep-learning approaches formulated super-resolution (SR) as deterministic regression. Methods such as SRCNN and EDSR~\cite{SRCNN,EDSR}, together with later architectures including RDN and RCAN~\cite{RDN2018,RCAN2018}, were trained using pixel-wise reconstruction losses. While effective at reducing distortion, these models often produce oversmoothed textures due to their limited ability to synthesise realistic high-frequency details.

To improve perceptual quality, subsequent work introduced stronger generative priors. GAN-based methods~\cite{SRGAN,ESRGAN} enhance perceptual realism but may suffer from training instability and hallucinated textures. More recently, transformer-based architectures such as SwinIR~\cite{SwinIR}, SwinFIR~\cite{SwinFIR}, and SRFormer~\cite{SRFormer}, together with degradation-aware pipelines like BSRGAN~\cite{BSRGAN}, have further improved robustness under complex real-world degradations~\cite{CSNL2020,IPT2021}.

Recent advances in generative modelling have also shifted attention toward diffusion-based formulations. Methods including SRDiff~\cite{SRDiff}, IDM~\cite{IDM}, ACDMSR~\cite{ACDMSR}, ResShift~\cite{ResShift}, StableSR~\cite{StableSR}, and related DDPM/DDIM-inspired frameworks demonstrate that iterative denoising can provide a stable pathway for generative reconstruction~\cite{FluxMapSR2601,TADiSR2506,FluxSR2505,RSDiffSR2025}.

In parallel, multiscale representations offer a principled framework for modelling hierarchical image structures. Wavelet and frequency-domain decompositions separate coarse and fine components while preserving spatial localisation, enabling access to cross-resolution statistics. Recent SR approaches operating in wavelet or frequency domains~\cite{DiWa,SinSR,FDDiff} suggest that frequency-aware processing can regularise texture synthesis and facilitate coarse-to-fine refinement.

Despite their strong generative capacity, most learning-based SR frameworks, including diffusion-based models, rely on large-scale external datasets with paired LR–HR supervision. This reliance makes multi-scale approaches difficult to apply in practice and ties performance closely to the availability and representativeness of training data. In contrast, SISR departs from this paradigm by exploiting internal image statistics within the input image itself.

\subsection{Single-Image Super-Resolution}

SISR adopts an image-adaptive learning model in which the reconstruction prior is inferred directly from internal statistics of the observed image. Rather than relying on externally learned mappings, this paradigm exploits the recurrence of local patterns and self-similarity inherent in natural images.

Early formulations instantiated this idea through explicit image-specific regression models. ZSSR~\cite{ZSSR} trained a lightweight predictor on patches extracted from the test image itself, demonstrating that internal examples can compensate for unknown degradations. Complementary approaches replaced explicit regression with implicit priors induced by network structure and test-time optimization, as in Deep Image Prior~\cite{DIP}. Extending this concept to generative modeling, SinGAN~\cite{SinGAN} learned a multiscale internal generator by matching patch distributions across resolutions.

More recently, diffusion-based models have been trained directly on a single image and sampled in a coarse-to-fine manner~\cite{SinDDM,SinFusion}, indicating that probabilistic generative priors can also be realized within an internal learning setting. Nevertheless, the LR observation remains inherently ambiguous. While internal learning can capture image-specific statistics, it does not explicitly encode structured dependencies across frequency levels, which are crucial for stabilizing high-frequency generation.






\section{Preliminaries}
\label{sec:prelim}

Motivated by the multiscale and cross-scale considerations, we formulate the reconstruction problem within a structured probabilistic framework.

\textbf{Degradation model.}
We observe a low-resolution (LR) image $y\in\mathbb{R}^{h\times w}$ generated from a high-resolution (HR) image $x\in\mathbb{R}^{H\times W}$ via
\begin{equation}
y = \mathcal{D}(x) + n,
\label{eq:degradation}
\end{equation}
where $\mathcal{D}(\cdot)$ denotes a degradation operator (e.g., blur, downsampling) and $n$ is additive noise. 
Since $x$ is unobserved in our formulation, the LR observation is enforced through an explicit LR-consistency constraint during inference. In practice, this update uses a differentiable instantiation of $\mathcal{D}$ consistent with the assumed degradation model in the experimental setting.

\textbf{DDPM recap.}
A denoising diffusion probabilistic model (DDPM)~\cite{ho2020ddpm} defines a forward noising process
\begin{equation}
q(x_t\mid x_0) = \mathcal{N}\!\left(x_t;\sqrt{\bar{\alpha}_t}\,x_0,(1-\bar{\alpha}_t)I\right),
\label{eq:ddpm_forward}
\end{equation}
with $\alpha_t=1-\beta_t$ and $\bar{\alpha}_t=\prod_{i=1}^t\alpha_i$. The reverse process is parameterized as Gaussian transitions
\begin{equation}
p_\theta(x_{t-1}\mid x_t)=\mathcal{N}\!\left(x_{t-1};\mu_\theta(x_t,t),\sigma_t^2 I\right),
\label{eq:ddpm_reverse}
\end{equation}
where $\mu_\theta(\cdot)$ is computed from a noise-prediction network $\epsilon_\theta$ using the standard DDPM parameterization.

\textbf{\`A trous multiscale representation.}
Let $x_{\mathrm{ref}}\in\mathbb{R}^{H\times W}$ denote an HR-grid reference defined on the target HR.
An \emph{undecimated} \`a trous transform produces a sequence of smooth components $\{c^{(s)}\}_{s=0}^S$ and detail planes $\{w^{(s)}\}_{s=1}^S$ on the HR grid:
\begin{equation}
c^{(0)}=x_{\mathrm{ref}},\qquad c^{(s)}=c^{(s-1)}*k^{(s)},\qquad
w^{(s)}=c^{(s-1)}-c^{(s)},
\label{eq:atrous}
\end{equation}
where $k^{(s)}$ denotes the dilated low-pass filter corresponding to the standard \`a trous B3-spline scaling function. The reconstruction identity is:
\begin{equation}
x_{\mathrm{ref}}=c^{(S)}+\sum_{s=1}^S w^{(s)}.
\label{eq:recon}
\end{equation}

Within this specified setting, the proposed reconstruction model can be derived.

\begin{figure*}[t]
  \centering
  \includegraphics[width=\textwidth]{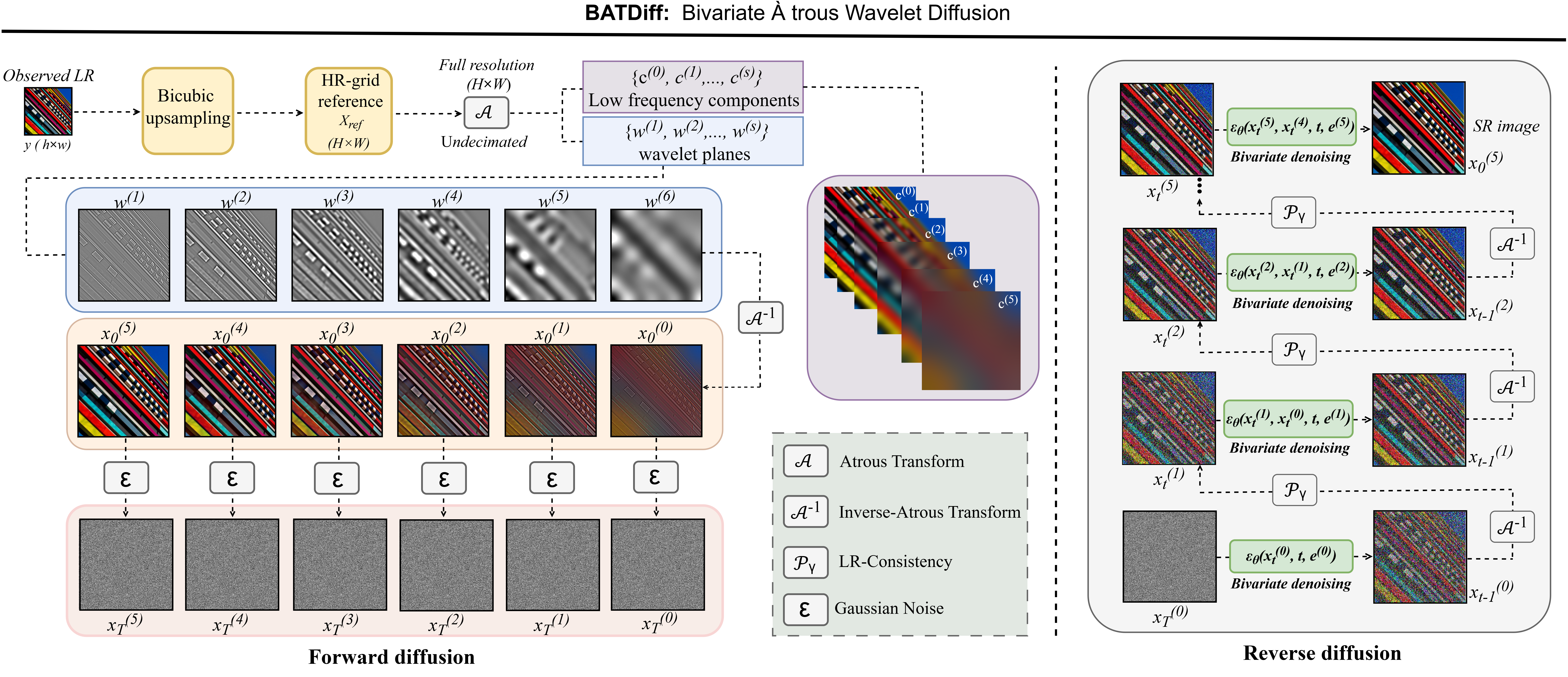}
  \caption{\textbf{Overview of BATDiff.}
Given an observed LR image $y$, we form an HR-grid reference $x_{\mathrm{ref}}=\mathcal{U}(y)$ via bicubic upsampling and build an undecimated \`a trous pyramid $\mathcal{A}$ to obtain multiscale wavelet planes $\{w^{(s)}\}$ (\textit{S}=6).
We construct progressively refined clean targets $\{x^{(s)}_{0}\}$ (Eq.~\eqref{eq:partial_method}) and train a shared DDPM by forward noising at each scale.
At inference, reverse diffusion proceeds from coarse to fine with \emph{bivariate} denoising conditioned on the time-aligned parent state $x^{(s-1)}_{t}$, followed by an LR-consistency correction $\mathcal{P}_y$ after each step.}
  \label{fig:pipeline}
\end{figure*}

\section{Proposed Method}
\label{sec:method}
Fig.~\ref{fig:pipeline} summarizes BATDiff. Given an LR observation $y$, we first construct an HR-grid reference $x_{\mathrm{ref}}=\mathcal{U}(y)$ using an upsampling operator $\mathcal{U}$. An \`a trous hierarchy (Sec.~\ref{sec:prelim}) is then built from $x_{\mathrm{ref}}$ to define scale-indexed clean targets $\{x^{(s)}_0\}_{s=0}^{S}$ that progressively reveal high-frequency content while preserving spatial alignment. We train a shared DDPM noise-predictor on these targets by synthetically noising them with the DDPM forward process.
During inference, given an LR observation $y$ (Eq.~\eqref{eq:degradation}), we sample sequentially from coarse to fine scales using the bivariate reverse transitions in Sec.~\ref{sec:bivar}, while enforcing LR-consistency via Eq.~\eqref{eq:lrcorr_method} after each reverse step.

\subsection{Diffusion variables}
\label{sec:scalevars}
From \`a trous decomposition (Eqs.~\eqref{eq:atrous}--\eqref{eq:recon}), we form the partial reconstructions
\begin{equation}
x^{(0)}_0 = c^{(S)}, \qquad
x^{(s)}_0 = c^{(S)} + \sum_{j=1}^{s} w^{(j)}, \quad s=1,\dots,S,
\label{eq:partial_method}
\end{equation}
which define a coarse-to-fine refinement path on a common spatial grid. For each scale $s$, we run the standard DDPM forward process (Eq.~\eqref{eq:ddpm_forward}) on $x^{(s)}_0$ to obtain noisy states $\{x^{(s)}_t\}_{t=1}^{T}$. The noise schedule $\{\beta_t\}$ and the forward dynamics are shared across scales; scale enters only through the clean target $x^{(s)}_0$.

\subsection{Bivariate reverse diffusion}
\label{sec:bivar}
\textbf{Cross-scale structure.}
A scale-separated construction would model each hierarchy level independently via
$p_\theta(x^{(s)}_{t-1}\mid x^{(s)}_{t})$, which leaves the fine-scale trajectory weakly constrained by the evolving coarse structure. Parent–child interscale dependencies are well documented in wavelet-domain restoration and have been exploited via bivariate estimators that condition a coefficient on its parent to improve stability and suppress artifacts~\cite{Alin}. BATDiff instead imposes explicit parent-child coupling across adjacent \`a trous scales by modeling, for $s=1,\dots,S$,
\begin{equation}
p_\theta\!\left(x^{(s)}_{t-1}\mid x^{(s)}_{t},\,x^{(s-1)}_{t}\right),
\label{eq:bivar_method}
\end{equation}
with the coarsest scale as the standard univariate special case
$p_\theta(x^{(0)}_{t-1}\mid x^{(0)}_{t})$.
The conditioning variable $x^{(s-1)}_t$ is time-aligned with $x^{(s)}_t$, providing a contemporaneous coarse reference that suppresses cross-scale drift. Algorithm~\ref{alg:batdiff} summarizes the sampling procedure.

We parameterize all reverse steps using a single noise-prediction network $\epsilon_\theta$ shared across scales. The scale index is provided through a learned embedding $\mathbf{e}(s)$, allowing the same network to represent scale-dependent denoising behaviour. Concretely,
\begin{align}
\hat{\epsilon}^{(0)}_t &= \epsilon_\theta\!\left(x^{(0)}_t,\, t,\, \mathbf{e}(0)\right),
\label{eq:eps0_method}\\
\hat{\epsilon}^{(s)}_t &= \epsilon_\theta\!\left(x^{(s)}_t,\, x^{(s-1)}_t,\, t,\, \mathbf{e}(s)\right), \qquad s=1,\dots,S.
\label{eq:eps_method}
\end{align}
This bivariate input is the defining structural element of BATDiff; the reverse diffusion parameterization remains the standard DDPM Gaussian form (Eq.~\eqref{eq:ddpm_reverse}).

\textbf{Reverse update}
Given $\hat{\epsilon}^{(s)}_t$, we compute the reverse mean using the standard DDPM noise-parameterization
\begin{equation}
\mu^{(s)}_\theta \;=\; \frac{1}{\sqrt{\alpha_t}}
\left(x^{(s)}_t - \frac{1-\alpha_t}{\sqrt{1-\bar{\alpha}_t}}\,\hat{\epsilon}^{(s)}_t\right),
\label{eq:mu_method}
\end{equation}
and sample $x^{(s)}_{t-1}\sim \mathcal{N}(\mu^{(s)}_\theta,\sigma_t^2 I)$. The parent $x^{(s-1)}_t$ influences the update only through $\hat{\epsilon}^{(s)}_t$ in Eq.~\eqref{eq:eps_method}.

\textbf{LR-consistency.}
We enforce fidelity to the LR observation $y$ by penalizing violations of the degradation model in Eq.~\eqref{eq:degradation}. Concretely, we use the data-consistency objective
\begin{equation}
\mathcal{L}_{\mathrm{lr}}(x)=\left\|\mathcal{D}(x)-y\right\|_2^2,
\label{eq:lrloss_method}
\end{equation}
where $\mathcal{D}$ is implemented as a differentiable degradation operator consistent with the assumed evaluation setting (e.g., blur+downsampling). After each reverse diffusion step at scale $s$ and timestep $t$ (i.e., after obtaining the intermediate iterate $x^{(s)}_{t-1}$), we apply a lightweight correction

\begin{equation}
x^{(s)}_{t-1} \leftarrow x^{(s)}_{t-1} - \eta_t \, \nabla_{x^{(s)}_{t-1}} \left\|\mathcal{D}\!\left(x^{(s)}_{t-1}\right)-y\right\|_2^2,
\label{eq:lrcorr_method}
\end{equation}
with step size $\eta_t$ kept constant across diffusion timesteps. Note that all $x^{(s)}_{t}$ live on the HR grid (undecimated \`a trous), hence $\mathcal{D}$ is applied consistently at every scale. When $\mathcal{D}$ admits a closed-form projection, Eq.~\eqref{eq:lrcorr_method} can be replaced by the corresponding projection operator. This design keeps the LR observation as an explicit inference-time constraint, while the diffusion prior supplies the missing high-frequency content.

\subsection{Training objective}
We optimize the standard DDPM noise-prediction objective, averaged over wavelet scales, diffusion timesteps, and Gaussian noise:
\begin{equation}
\mathcal{L}(\theta)
=
\mathbb{E}_{\substack{s\sim\mathrm{Unif}\{0,\dots,S\}\\
t\sim\mathrm{Unif}\{1,\dots,T\}\\
\epsilon\sim\mathcal{N}(0,I)}}
\left[
\left\|\epsilon-\hat{\epsilon}^{(s)}_t\right\|_2^2
\right].
\label{eq:loss_method}
\end{equation}
For a given scale $s$, the clean target $x^{(s)}_0$ is constructed deterministically from the reference image $x_{\mathrm{ref}}$ via Eq.~\eqref{eq:partial_method}. A noisy sample is then generated using the forward diffusion process (Eq.~\eqref{eq:ddpm_forward}),
\begin{equation}
x^{(s)}_t=\sqrt{\bar{\alpha}_t}\,x^{(s)}_0+\sqrt{1-\bar{\alpha}_t}\,\epsilon.
\end{equation}
The predicted noise $\hat{\epsilon}^{(s)}_t$ is given by Eq.~\eqref{eq:eps0_method} for $s=0$ and Eq.~\eqref{eq:eps_method} for $s\ge 1$. No paired LR--HR training set is used; the LR observation $y$ is used only at inference time via the LR-consistency updates (Eq.~\eqref{eq:lrcorr_method}).
\begin{algorithm}[t]
\caption{BATDiff inference with bivariate reverse diffusion and LR consistency}
\label{alg:batdiff}
\scriptsize
\DontPrintSemicolon
\KwIn{LR $y$, scales $S$, steps $T$, noise schedule $\{\beta_t\}$, denoiser $\epsilon_\theta$, degradation operator $\mathcal{D}$, step sizes $\{\eta_t\}$}
\KwOut{HR estimate $\hat{x}$}

Sample $x_T^{(0)} \sim \mathcal{N}(0,I)$\;

\For{$s=0$ \KwTo $S$}{
    \If{$s>0$}{
        $x_T^{(s)} \leftarrow \hat{x}_0^{(s-1)}$\;
    }
    \For{$t=T, T-1, ...,1$}{
        \uIf{$s=0$}{
            $\hat{\epsilon}_t^{(0)} \leftarrow \epsilon_\theta(x_t^{(0)}, t, \mathbf{e}(0))$\;
        }\Else{
            $\hat{\epsilon}_t^{(s)} \leftarrow \epsilon_\theta(x_t^{(s)}, x_t^{(s-1)}, t, \mathbf{e}(s))$\;
        }
        $\mu_\theta^{(s)} \leftarrow \frac{1}{\sqrt{\alpha_t}}
        \bigl(x_t^{(s)} - \frac{1-\alpha_t}{\sqrt{1-\bar{\alpha}_t}}\hat{\epsilon}_t^{(s)}\bigr)$\;
        Sample $x_{t-1}^{(s)} \sim \mathcal{N}(\mu_\theta^{(s)}, \sigma_t^2 I)$\;
        $x_{t-1}^{(s)} \leftarrow x_{t-1}^{(s)} - \eta_t \nabla_x \|\mathcal{D}(x_{t-1}^{(s)}) - y\|_2^2$\;
    }
    $\hat{x}_0^{(s)} \leftarrow x_0^{(s)}$\;
}
\Return $\hat{x} \leftarrow \hat{x}_0^{(S)}$\;
\end{algorithm}

\FloatBarrier
\section{Experiments}

\subsection{Experiment Setup}

\textbf{Evaluation benchmarks and protocol.}
We evaluate BATDiff on four standard SR datasets, namely DIV2K~\cite{DIV2K}, Set5~\cite{Set5}, Set14~\cite{Set14}, and Urban100~\cite{Urban100}. Bicubic interpolation is used as the upsampling operator $\mathcal{U}$. Following standard SR evaluation practice, PSNR and SSIM are computed. LPIPS is additionally reported in the ablation study to assess perceptual quality. Since BATDiff operates in a single-image/internal-learning regime, no external paired LR--HR training set is used. Instead, for each test image, scale-specific training samples are constructed from the image itself.

\textbf{Compared methods.}
We compare BATDiff with representative methods from several SR families, GAN-based, transformers and diffusion-based approaches. Specifically, the comparison includes ZSSR~\cite{ZSSR}, ESRGAN~\cite{ESRGAN}, BSRGAN~\cite{BSRGAN}, SwinIR~\cite{SwinIR}, SwinFIR~\cite{SwinFIR}, SRFormer~\cite{SRFormer}, SRDiff~\cite{SRDiff}, DiWa~\cite{DiWa}, FDDiff~\cite{FDDiff}, ResShift~\cite{ResShift}, StableSR~\cite{StableSR}, and SinSR~\cite{SinSR}.

\textbf{Implementation details.}
The model is optimized using Adam with an initial learning rate of $10^{-3}$ and a training batch size of 16. Training is performed for 120k training steps using a multi-step learning-rate schedule and the diffusion process uses 100 timesteps. For multiscale decomposition, we adopt a 6-level \`a trous pyramid. All experiments are conducted on an NVIDIA H100 GPU.

\subsection{Quantitative Comparison}
Table~\ref{tab:main_sr} reports quantitative comparisons on four standard SR benchmarks with the upsampling scale of 4. BATDiff operates in a single-image internal-learning regime without external paired LR--HR supervision. Only a limited number of SR methods are designed for this setting; therefore, ZSSR~\cite{ZSSR} serves as the most directly comparable baseline, while pretrained supervised models trained on large external datasets are reported mainly for contextual reference. On DIV2K ($\times4$), BATDiff obtains 29.34 dB PSNR and 0.8948 SSIM, achieving the strong performance. On Set5, it achieves 32.89 dB PSNR and the highest SSIM of 0.9063 and the best results in both PSNR and SSIM on Set14 with 30.12 dB and 0.8134, respectively. On Urban100, It again achieves the best performance with 28.53 dB PSNR and 0.8502 SSIM. BATDiff also demonstrates consistent performance at the more challenging $\times8$ scale factor across all datasets (as shown in Table~\ref{tab:main_sr_x8}), whereas supervised methods often fail to generalise to upsampling factors beyond the training distribution. These methods are trained on paired $\times4$ images and perform upsampling using coordinate-based representations, which provide significantly better reconstruction quality than bilinear or bicubic interpolation. In addition, since reconstruction is performed on an HR-grid reference rather than scale-specific upsampling layers, the proposed framework can naturally support non-integer magnification factors without architectural modification.

\begin{table*}[t]
\centering
\caption{Quantitative comparison on standard SR benchmarks with an up-sampling scale of $\times4$. \textbf{P}: paired, \textbf{U}: unpaired, \textbf{SI}: single-image. Best/second-best within each comparable block are \best{bold}/\second{underlined}.}

\label{tab:main_sr}
\footnotesize
\setlength{\tabcolsep}{2.4pt}
\renewcommand{\arraystretch}{0.88}
\begin{tabular}{@{}l l c >{\raggedright\arraybackslash}p{0.38\textwidth} c c@{}}
\toprule
Dataset & Method & Sup. & Training dataset & PSNR$\uparrow$ & SSIM$\uparrow$ \\
\midrule
\multirow{13}{*}[-0.1 em]{DIV2K}
& EDSR~\cite{EDSR}  & P  & DF2K                   & \second{28.98} & 0.83 \\
& BSRGAN~\cite{BSRGAN}  & P  & DIV2K+Flickr2K+WED+FFHQ & 22.56 & 0.7352 \\
\cmidrule(l){2-6}
& SRDiff~\cite{SRDiff}  & P  & DIV2K                   & 27.41 & 0.79 \\
& DiWa~\cite{DiWa}      & P  & DIV2K                   & 28.09 & 0.78 \\
& FDDiff~\cite{FDDiff}  & P  & DIV2K                   & 27.53 & \second{0.85} \\
\cmidrule(l){2-6}

& StableSR~\cite{StableSR} & P  & DIV2K+DIV8K+Flickr2K+OST     & 20.59 & 0.6562 \\
& SinSR~\cite{SinSR}       & P & ImageNet                     & 23.12 & 0.77 \\
\cmidrule(l){2-6}
& \textbf{Ours$\times4$}  & U/SI & SI                         & \best{29.34} & \best{0.8948} \\
\midrule

\multirow{7}{*}[-1.0 em]{Set5}
& ZSSR~\cite{ZSSR}        & U/SI & SI                 &  31.13 &  0.8796 \\
& BSRGAN~\cite{BSRGAN}      & P  & DIV2K+Flickr2K+WED+FFHQ & 21.98 & 0.8175 \\
\cmidrule(l){2-6}
& SwinIR~\cite{SwinIR}      & P  & DIV2K                   & 32.72 & 0.9021 \\
& SwinFIR~\cite{SwinFIR}    & P  & DF2K                    & \best{33.08} & \second{0.9048} \\
\cmidrule(l){2-6}
& StableSR~\cite{StableSR}  & P  & DIV2K+DIV8K+Flickr2K+OST & 20.35 & 0.7665 \\
& SinSR~\cite{SinSR}        & P & ImageNet                 & 23.91 & 0.8717 \\
\cmidrule(l){2-6}
& \textbf{Ours$\times4$}   & U/SI & SI                     & \second{32.89} & \best{0.9063} \\
\midrule

\multirow{7}{*}[-1.0 em]{Set14}
& ZSSR~\cite{ZSSR}        & U/SI & SI                 &  28.01 &  0.7651 \\
& BSRGAN~\cite{BSRGAN}      & P  & DIV2K+Flickr2K+WED+FFHQ & 29.41 & 0.7141 \\
\cmidrule(l){2-6}
& SwinIR~\cite{SwinIR}      & P  & DIV2K                   & 28.94 & 0.7914 \\
& SRFormer~\cite{SRFormer}  & P  & DF2K                    & 29.01 & \second{0.7919} \\
\cmidrule(l){2-6}
& StableSR~\cite{StableSR}  & P  & DIV2K+DIV8K+Flickr2K+OST & 28.17 & 0.7112 \\
& SinSR~\cite{SinSR}        & P & ImageNet                 & \second{29.47} & 0.7360 \\
\cmidrule(l){2-6}
& \textbf{Ours$\times4$}   & U/SI & SI                     & \best{30.12} & \best{0.8134} \\
\midrule

\multirow{7}{*}[-0.7 em]{Urban100}
& BSRGAN~\cite{BSRGAN}      & P  & DIV2K+Flickr2K+WED+FFHQ & 24.54 & 0.6426 \\
\cmidrule(l){2-6}
& SwinIR~\cite{SwinIR}      & P  & DIV2K                   & 27.07 & 0.8164 \\
& SRFormer~\cite{SRFormer}  & P  & DF2K                    & \second{27.85} & \second{0.8338} \\
\cmidrule(l){2-6}

& StableSR~\cite{StableSR}  & P  & DIV2K+DIV8K+Flickr2K+OST & 22.11 & 0.5423 \\
& SinSR~\cite{SinSR}        & P & ImageNet                 & 23.84 & 0.5964 \\
\cmidrule(l){2-6}
& \textbf{Ours$\times4$}   & U/SI & SI                     & \best{28.53} & \best{0.8502} \\
\bottomrule
\end{tabular}
\end{table*}

\begin{table}[t]
\centering
\caption{Quantitative comparison (PSNR$\uparrow$) on standard SR benchmarks with an up-sampling scale of $\times8$). Best/second-best within each comparable block are \best{bold}/\second{underlined}.}

\label{tab:main_sr_x8}
\footnotesize
\setlength{\tabcolsep}{2.4pt}
\renewcommand{\arraystretch}{0.88}
\begin{tabular}{l| ccccccc }
\toprule
Dataset & MetaSR~\cite{MetaSR} & LIIF~\cite{LIIF} & LTE~\cite{LTE}  & CLIT~\cite{CLIT} & CiaoSR~\cite{CiaoSR} & HIIF~\cite{HIIF} & Ours$\times8$ \\
\midrule

Set5 & 27.02 & 27.36 & 27.35 & \second{27.62} & 27.45 & 27.56 & \best{28.74}\\

Set14 & 25.09 & 25.34 & 25.42 & 25.55 & 25.42 & \second{25.59} & \best{27.12} \\

Urban100 &22.75 & 23.14 & 23.17 & 23.33 & 23.34 & \second{23.36}  & \best{23.47} \\

\bottomrule
\end{tabular}
\end{table}

\subsection{Qualitative Comparison}

\begin{figure*}[t]
  \centering
  \includegraphics[width=0.94\textwidth]{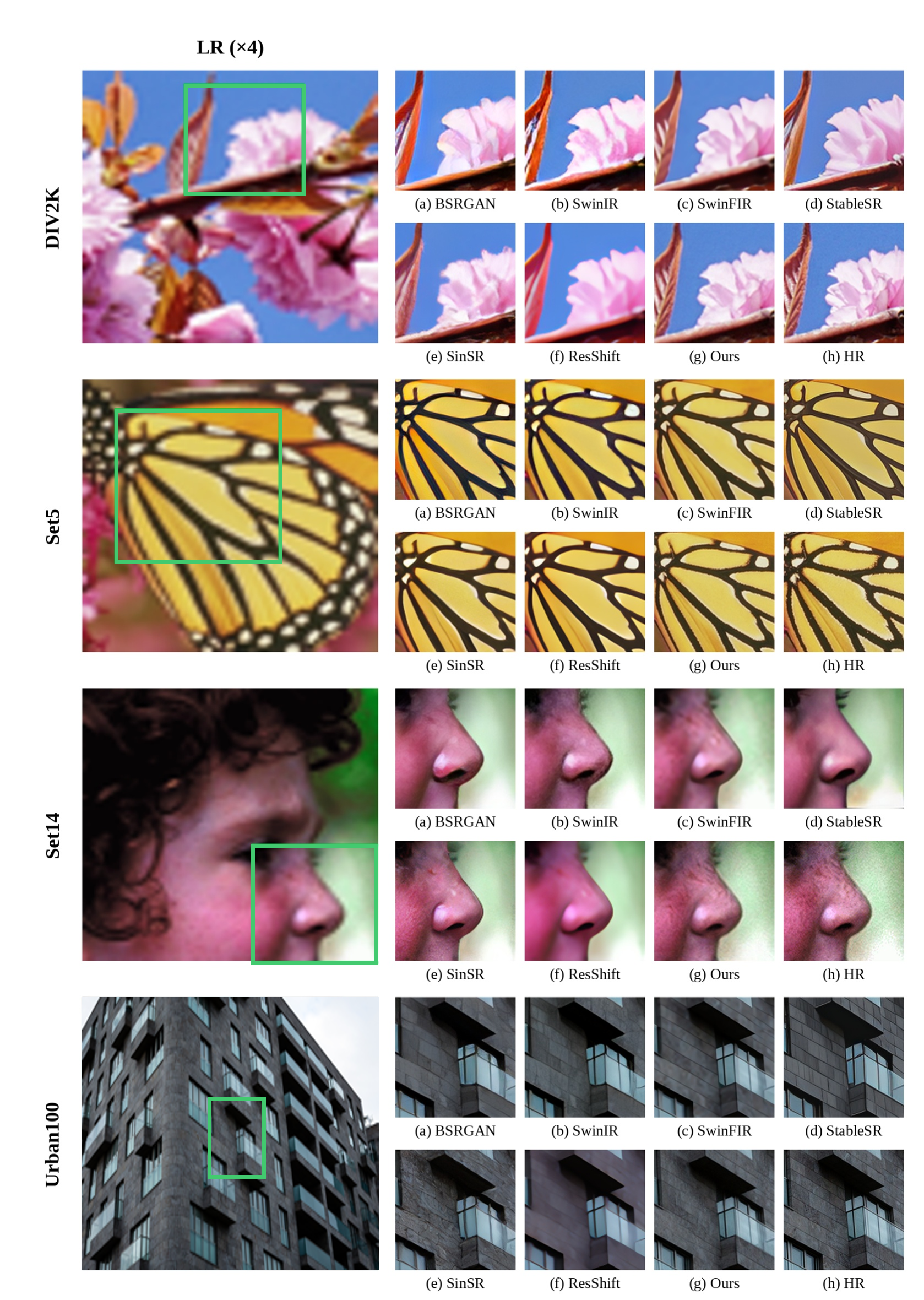}
  \caption{Qualitative comparison on standard SR benchmarks ($\times4$).}
  \label{fig:q1}
\end{figure*}

Representative visual comparisons are shown in Fig.~\ref{fig:q1}. BATDiff tends to reconstructs cleaner edges, sharper contours, and more plausible fine structures than other approaches. Improvements are especially evident in regions containing repeated textures, thin line patterns, and high-contrast boundaries, where regression-based methods tend to oversmooth details and generative models may introduce visually inconsistent artifacts. Compared with existing diffusion and non-diffusion methods, BATDiff produces reconstructions that are visually closer to the HR reference.

\subsection{Ablation Study}

\begin{figure}[htbp]
  \centering
  \includegraphics[width=0.55\columnwidth]{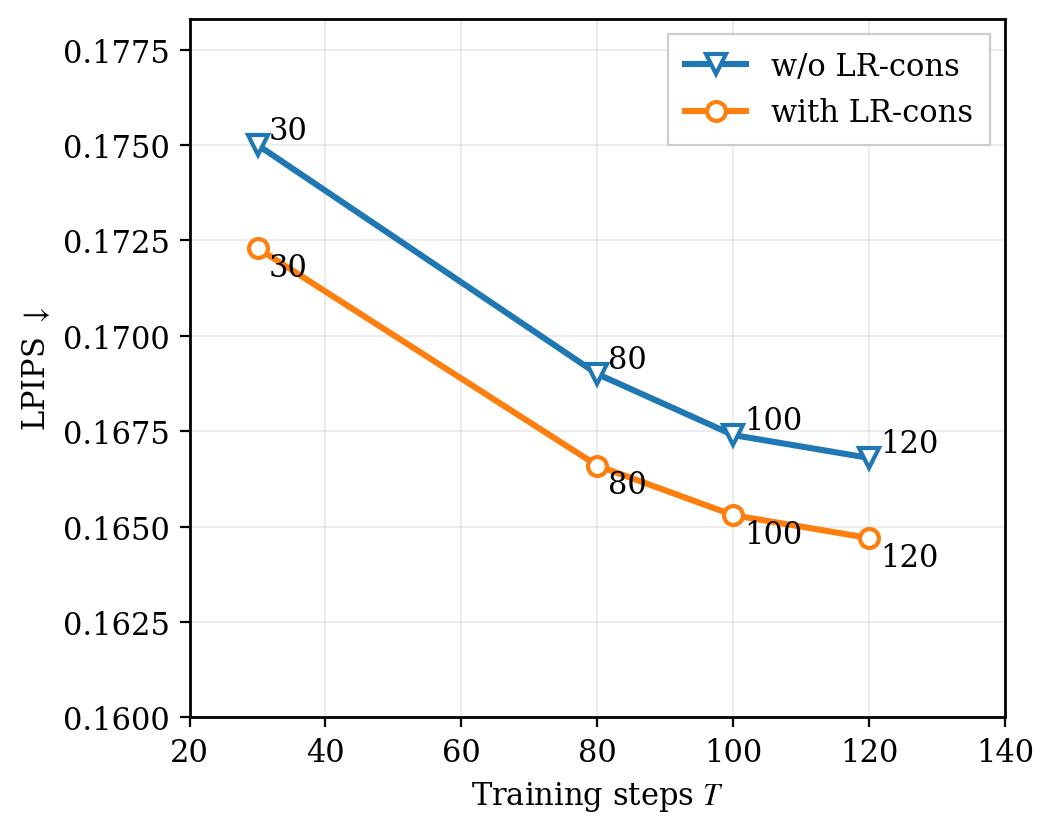}
  \caption{Effect of LR-consistency across training steps.}
  \label{fig:q4}
\end{figure}

\begin{figure*}[!b]
  \centering
  \includegraphics[width=0.95\textwidth]{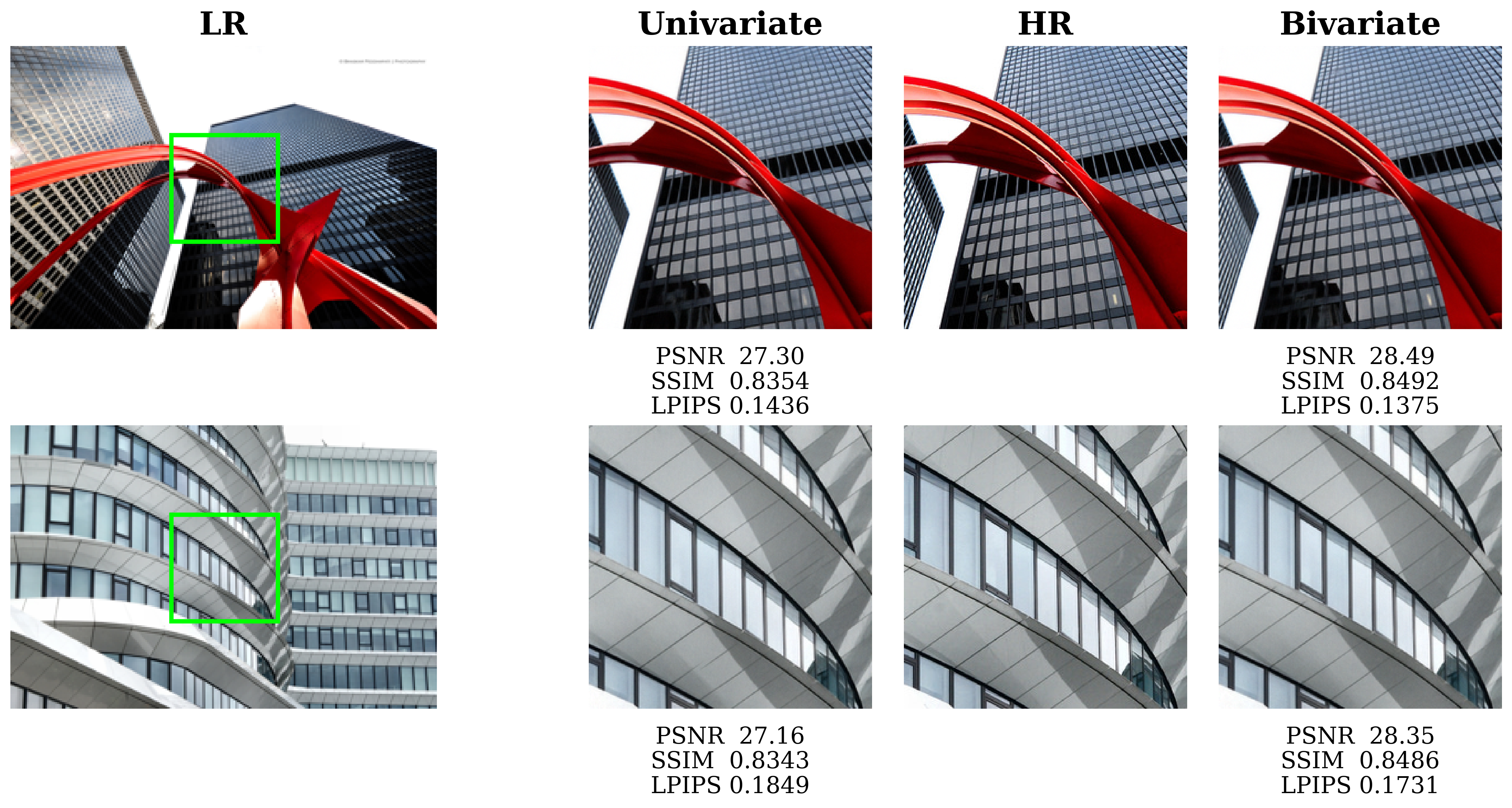}
  \caption{Qualitative ablation on Urban100 ($\times4$). LR, univariate, HR, and BATDiff results are shown.}
  \label{fig:q2}
\end{figure*}

\begin{figure*}[t]
  \centering
  \includegraphics[width=0.95\textwidth]{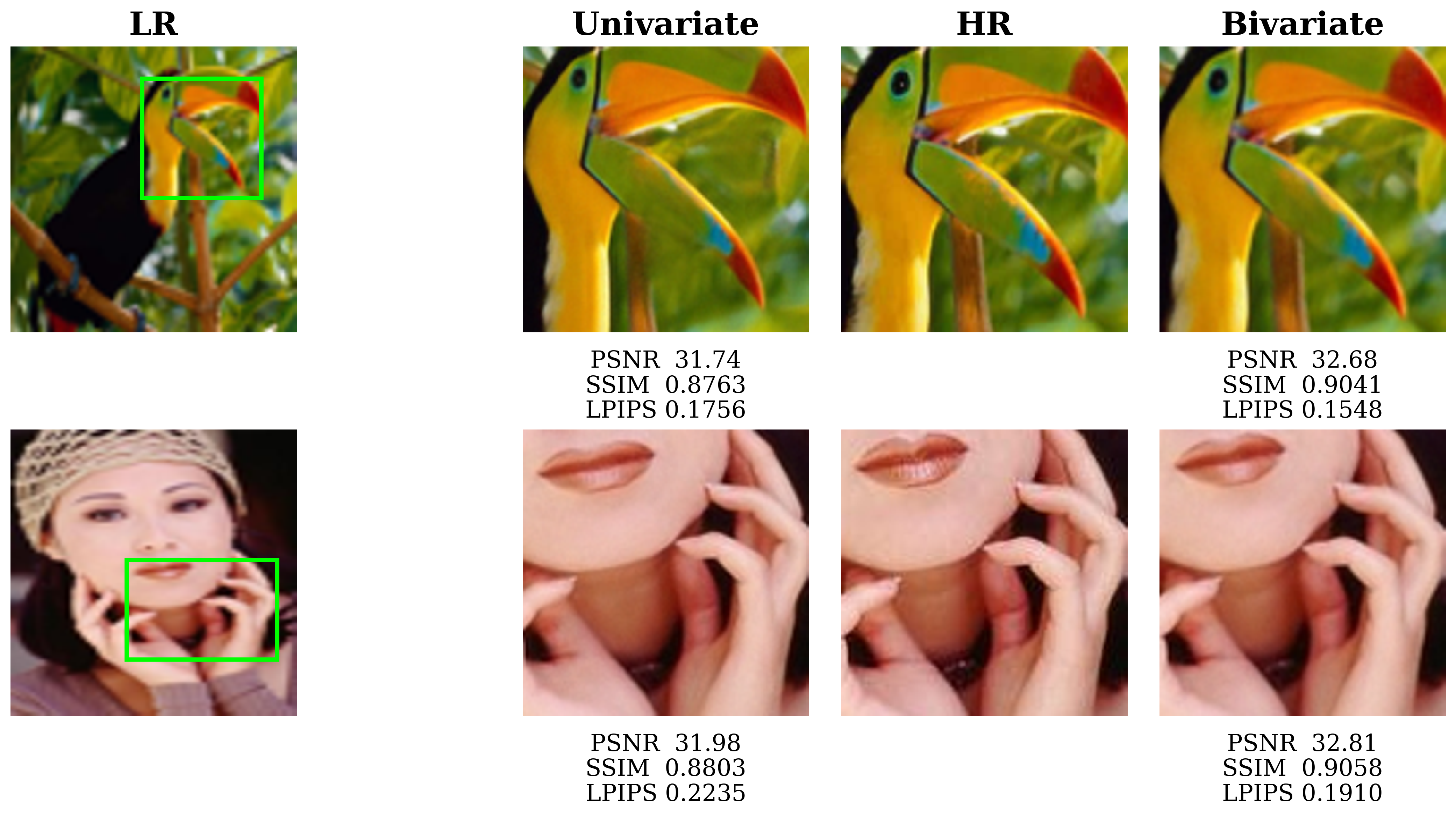}
  \caption{Qualitative ablation on Set14 ($\times4$). LR, univariate, HR, and BATDiff results.}
  \label{fig:q3}
\end{figure*}

The contribution of the main components is analyzed on Urban100 under the $\times4$ Bicubic setting. Quantitative results are summarized in Tables~\ref{tab:ablation_core}--\ref{tab:sensitivity_T_w_d_transposed}, while qualitative examples are shown in Figs.~\ref{fig:q4}--\ref{fig:q3}. Urban100 is particularly suitable for ablation because it contains strong geometric regularity and dense repeated high-frequency patterns.

\textbf{Effect of core components.}
Table~\ref{tab:ablation_core} evaluates the cumulative contribution of LR-consistency, the \`a trous multiscale representation, and the proposed bivariate reverse diffusion. Using LR-consistency alone yields 26.93 dB PSNR, 0.8240 SSIM, and 0.1784 LPIPS. Introducing the \`a trous decomposition improves all metrics, increasing PSNR to 27.46 dB and SSIM to 0.8377 while reducing LPIPS to 0.1723. When the proposed bivariate conditioning is further incorporated, performance improves substantially to 28.53 dB PSNR, 0.8502 SSIM, and 0.1647 LPIPS. Also, Fig.~\ref{fig:q4} illustrates the effect of LR-consistency across training steps. Enforcing LR-consistency improves the sampling trajectory throughout the reverse process and yields lower perceptual error as the number of steps increases. 

\begin{table}[t]
\centering
\caption{Ablation of core components on Urban100 ($\times4$). All variants use the same backbone and training steps ($T{=}120$).}
\label{tab:ablation_core}
\footnotesize
\setlength{\tabcolsep}{3.2pt}
\renewcommand{\arraystretch}{0.94}
\begin{tabular}{ccc|ccc}
\toprule
LR-cons. & \`A trous & Bivariate & PSNR$\uparrow$ & SSIM$\uparrow$ & LPIPS$\downarrow$ \\
\midrule
\cmark & \xmark & \xmark & 26.93 & 0.8240 & 0.1784 \\
\cmark & \cmark & \xmark & 27.46 & 0.8377 & 0.1723 \\
\cmark & \cmark & \cmark & \best{28.53} & \best{0.8502} & \best{0.1647} \\
\bottomrule
\end{tabular}
\vspace{4mm}
\centering
\caption{Ablation of parent choice in the bivariate design on Urban100 ($\times4$).}
\label{tab:ablation_bivar}
\footnotesize
\setlength{\tabcolsep}{3.4pt}
\renewcommand{\arraystretch}{0.95}
\begin{tabular}{lccc}
\toprule
Parent in Eq.~\eqref{eq:bivar_method} & PSNR$\uparrow$ & SSIM$\uparrow$ & LPIPS$\downarrow$ \\
\midrule
None (univariate)            & 27.46 & 0.8377 & 0.1723 \\
Misaligned $x^{(s-1)}_{t-1}$ & 27.92 & 0.8426 & 0.1704 \\
Coarse final $x^{(s-1)}_{0}$ & 28.14 & 0.8450 & 0.1668 \\
Time-aligned $x^{(s-1)}_{t}$ & \best{28.53} & \best{0.8502} & \best{0.1647} \\
\bottomrule
\end{tabular}
\end{table}

\textbf{Effect of the parent signal.}
Table~\ref{tab:ablation_bivar} studies different choices for the parent representation used in the bivariate reverse process. Removing parent conditioning entirely reduces the model to an univariate variant and yields the weakest results. Replacing the parent with a temporally misaligned state $x^{(s-1)}_{t-1}$ yields only a modest improvement, showing that cross-scale context is useful but sensitive to temporal inconsistency. Using the final coarse estimate $x^{(s-1)}_{0}$ performs better, as it provides a more stable guide. However, the best performance is achieved when the denoiser is conditioned on the time-aligned parent state $x^{(s-1)}_{t}$.

This comparison clarifies that the advantage of BATDiff is not merely due to injecting additional information into the network. Rather, the gain depends specifically on preserving temporal alignment across adjacent scales during sampling. The time-aligned parent acts as a contemporaneous structural prior, reducing cross-scale drift and improving the stability of fine-detail generation.

\textbf{Sensitivity to training steps and hyperparameters.}
Table~\ref{tab:sensitivity_T_w_d_transposed} analyzes sensitivity to the training steps $T$ and the hyperparameters $\omega$ and $d$. Here, $\omega$ controls the posterior noise level during reverse diffusion, while $d$ denotes the detail-gain coefficient in the \`a trous branch which controls the strength of wavelet-detail injection during multiscale reconstruction. Increasing the training budget from 30 to 120 steps consistently improves all metrics, with the best performance obtained at $T=120$. We further observe that the best operating point is obtained at $\omega=0.3$ and $d=0.8$. Increasing $\omega$ beyond this value degrades both fidelity and perceptual quality, suggesting that excessively strong correction perturbs the generative trajectory. Similarly, moving $d$ away from the selected setting leads to a gradual reduction in performance.

\begin{table}[t]
\centering
\caption{Sensitivity to training steps and hyperparameters on Urban100 ($\times4$).}
\label{tab:sensitivity_T_w_d_transposed}
\footnotesize
\setlength{\tabcolsep}{3pt}
\renewcommand{\arraystretch}{0.94}
\resizebox{\columnwidth}{!}{%
\begin{tabular}{l|cccc|cccc|cccc}
\toprule
 & \multicolumn{4}{c|}{$T$ sweep} & \multicolumn{4}{c|}{$\omega$ sweep} & \multicolumn{4}{c}{$d$ sweep} \\
\midrule
$T$      & 30  & 80  & 100 & 120 & 120 & 120 & 120 & 120 & 120 & 120 & 120 & 120 \\
$\omega$ & 0.3 & 0.3 & 0.3 & 0.3 & 0.3 & 0.5 & 1.0 & 2.0 & 0.3 & 0.3 & 0.3 & 0.3 \\
$d$      & 0.8 & 0.8 & 0.8 & 0.8 & 0.8 & 0.8 & 0.8 & 0.8 & 0.5 & 0.8 & 1.0 & 1.5 \\
\midrule
PSNR$\uparrow$  & 28.05 & 28.42 & 28.49 & \best{28.53} & \best{28.53} & 28.32 & 28.38 & 28.12 & 28.50 & \best{28.53} & 28.47 & 28.22 \\
SSIM$\uparrow$  & 0.8440 & 0.8485 & 0.8494 & \best{0.8502} & \best{0.8502} & 0.8491 & 0.8488 & 0.8450 & 0.8495 & \best{0.8502} & 0.8427 & 0.8416 \\
LPIPS$\downarrow$ & 0.1759 & 0.1697 & 0.1663 & \best{0.1647} & \best{0.1647} & 0.1654 & 0.1710 & 0.1769 & 0.1685 & \best{0.1647} & 0.1732 & 0.1761 \\
\bottomrule
\end{tabular}
}
\end{table}

\textbf{Univariate vs. bivariate models}
Figs.~\ref{fig:q2} and \ref{fig:q3} compare the univariate baseline with the proposed bivariate model on representative examples from Urban100 and Set14. In both cases, the bivariate formulation reconstructs sharper boundaries and visually cleaner local textures. 

Beyond visual sharpness, the bivariate variant exhibits improved structural continuity along elongated edges and repeated line patterns, with fewer local phase inconsistencies. Compared to the univariate baseline, it reduces local structural jitter and discontinuities that typically emerge during fine-scale denoising.  These observations are consistent with the quantitative gains in SSIM and LPIPS, reflecting better alignment between synthesized details and the underlying coarse structure.

\FloatBarrier
\section{Conclusion}

This paper proposed BATDiff, an unsupervised diffusion-based framework for single-image super-resolution(SISR) that addresses a structural limitation in most existing diffusion-based SR frameworks: high-frequency details are often reconstructed without an explicit mechanism for preserving structural coherence during sampling. BATDiff addresses this limitation through a progressive multiscale reconstruction strategy, where fine details are recovered step by step under the guidance of structurally aligned coarser-scale information. In parallel, the reconstruction is constrained throughout the inference to remain consistent with the observed low-resolution input, rather than relying solely on the generative prior. The experimental results on standard benchmarks showed that this formulation yields strong performance and improved structural consistency relative to the representative baselines. The ablation results further suggested that the performance gains are associated with the combination of progressive multiscale reconstruction, cross-scale guidance, and inference-time observation consistency. Overall, the results indicate that imposing cross-scale structure directly within reconstruction provides a promising direction for incorporating multiscale dependencies in diffusion-based SISR.

\clearpage
\appendix
\section*{Appendix}
\addcontentsline{toc}{section}{Appendix}
\section{Implementation Details}
\label{sec:supp_impl}

For completeness and reproducibility, this section provides additional implementation details. In particular, we describe the internal learning setting, the construction of the multiscale training data, the optimization protocol, and the inference configuration used throughout the experimental evaluation.

\subsection{Training Setup}

BATDiff is trained under a single-image internal learning paradigm. For each test image, the model is optimized directly on samples constructed from the image itself, rather than relying on an external paired LR--HR training set. An HR-grid reference image is first obtained via bicubic upsampling of the observed LR input, after which an undecimated \`a trous decomposition~\cite{Atrous} is applied in order to generate the scale-wise representations used during training.

This multiscale construction allows the model to learn reconstruction dynamics across progressively finer frequency bands while preserving spatial correspondence between scales. The complete set of implementation choices adopted across all experiments is summarized in Table~\ref{tab:impl_details}.

\begin{table}[h]
\centering
\caption{Implementation details used in all experiments.}
\label{tab:impl_details}
\footnotesize
\setlength{\tabcolsep}{5pt}
\renewcommand{\arraystretch}{0.95}
\begin{tabular}{lc}
\toprule
Parameter & Value \\
\midrule
Optimizer & Adam \\
Initial learning rate & $1\times10^{-3}$ \\
Batch size & 16 \\
Training iterations & 120k \\
Diffusion timesteps & 100 \\
Number of \`a trous levels & 6 \\
Upsampling initialization & Bicubic \\
GPU & NVIDIA H100 \\
\bottomrule
\end{tabular}
\end{table}

\subsection{Inference Details}

At inference time, reconstruction proceeds sequentially from coarse to fine wavelet scales. At each scale, reverse diffusion refines the current estimate by sampling from the posterior distribution conditioned on the representation at the immediately coarser scale, ensuring that fine-scale details remain consistent with the previously reconstructed structure.

Following each reverse diffusion step, the intermediate reconstruction is further corrected with the LR-consistency update described in the main paper. This update enforces consistency with the observed LR input while allowing the diffusion prior to recover high-frequency structures that are not explicitly present in the degraded observation. The influence of several inference-related design choices is further examined in Tables~\ref{tab:eta_ablation} and \ref{tab:steps_ablation}.

\subsection{Shared Denoiser}

BATDiff employs a single denoising network that is shared across all wavelet scales. Rather than training independent denoisers for different scales, the model conditions the shared network on the current scale index through a learned scale embedding, enabling the same denoiser to adapt its behaviour according to the frequency level being reconstructed.

This design substantially improves parameter efficiency and avoids the need to optimize multiple models in a scale-adaptive fashion. A direct comparison between the shared denoiser design and a scale-specific alternative is reported in Table~\ref{tab:shared_vs_sep}.

\FloatBarrier
\section{Ablation Studies}
\label{sec:supp_ablation}

This section presents additional ablation experiments that complement the analyses reported in the main paper. Unless otherwise specified, all results are reported on Urban100~\cite{Urban100} under the $\times4$ super-resolution setting.

\subsection{Effect of the Number of Wavelet Levels}

We first examine the influence of the number of \`a trous decomposition levels used in the multiscale hierarchy. As shown in Table~\ref{tab:scales_ablation}, increasing the number of scales improves reconstruction performance up to six levels, beyond which the gain becomes marginal. We observe similar behavior for larger magnification factors ($\times6$ and $\times8$), and therefore adopt six wavelet levels as the default configuration throughout the experiments.

\begin{table}[h]
\centering
\caption{Effect of the number of \`a trous levels on Urban100 ($\times4$).}
\label{tab:scales_ablation}
\footnotesize
\setlength{\tabcolsep}{5pt}
\renewcommand{\arraystretch}{0.95}
\begin{tabular}{cccc}
\toprule
Scales & PSNR$\uparrow$ & SSIM$\uparrow$ & LPIPS$\downarrow$ \\
\midrule
4 & 27.92 & 0.8421 & 0.1712 \\
5 & 28.21 & 0.8465 & 0.1680 \\
6 & \best{28.53} & \best{0.8502} & \best{0.1647} \\
7 & 28.49 & 0.8496 & 0.1651 \\
\bottomrule
\end{tabular}
\end{table}

These results suggest that shallow pyramids do not expose sufficient cross-scale structure, whereas deeper hierarchies offer only limited additional benefit beyond the selected configuration.

\subsection{Effect of LR-Consistency Strength}

We further investigate the influence of the LR-consistency correction strength. Recall that, after each reverse diffusion step, the intermediate estimate is updated according to
\begin{equation}
x^{(s)}_{t-1} \leftarrow x^{(s)}_{t-1} - \eta \,\nabla_{x^{(s)}_{t-1}}
\left\|\mathcal{D}\!\left(x^{(s)}_{t-1}\right)-y\right\|_2^2,
\label{eq:supp_eta_update}
\end{equation}
where $\eta$ controls the strength of the correction toward consistency with the observed LR input. The quantitative results reported in Table~\ref{tab:eta_ablation} show that a moderate correction strength provides the most favorable balance between fidelity to the LR observation and generative flexibility.

\begin{table}[h]
\centering
\caption{Effect of LR-consistency step size on Urban100 ($\times4$).}
\label{tab:eta_ablation}
\footnotesize
\setlength{\tabcolsep}{5pt}
\renewcommand{\arraystretch}{0.95}
\begin{tabular}{cccc}
\toprule
$\eta$ & PSNR$\uparrow$ & SSIM$\uparrow$ & LPIPS$\downarrow$ \\
\midrule
0.1 & 28.12 & 0.8467 & 0.1689 \\
0.3 & \best{28.53} & \best{0.8502} & \best{0.1647} \\
0.5 & 28.28 & 0.8489 & 0.1668 \\
\bottomrule
\end{tabular}
\end{table}

When the correction is too weak, the reconstruction remains insufficiently constrained by the observed LR input. Conversely, an overly strong correction perturbs the generative trajectory and leads to a modest degradation in reconstruction quality.

\subsection{Shared vs.\ Scale-Specific Denoisers}

We compare the proposed shared denoiser formulation with a variant that trains an independent denoising network for each scale. As shown in Table~\ref{tab:shared_vs_sep}, the shared design achieves slightly better reconstruction performance while being substantially more parameter efficient.

\begin{table}[h]
\centering
\caption{Shared vs.\ scale-specific denoisers on Urban100 ($\times4$).}
\label{tab:shared_vs_sep}
\footnotesize
\setlength{\tabcolsep}{5pt}
\renewcommand{\arraystretch}{0.95}
\begin{tabular}{lccc}
\toprule
Model & PSNR$\uparrow$ & SSIM$\uparrow$ & LPIPS$\downarrow$ \\
\midrule
Separate network per scale & 28.41 & 0.8485 & 0.1669 \\
Shared network (ours) & \best{28.53} & \best{0.8502} & \best{0.1647} \\
\bottomrule
\end{tabular}
\end{table}

This result indicates that exposing a single denoiser to multiple scales is beneficial, likely because it promotes parameter sharing across structurally related reconstruction stages.

\subsection{Effect of Reverse Diffusion Steps}

This experiment examines the trade-off between the number of reverse diffusion steps and reconstruction quality. The corresponding evaluation results are summarized in Table~\ref{tab:steps_ablation}. As the number of reverse diffusion steps increases, reconstruction quality gradually improves. The best performance is achieved with 100 steps, suggesting that this setting provides a good balance between reconstruction accuracy and computational efficiency, while further increasing the number of steps yields only marginal gains.

\begin{table}[h]
\centering
\caption{Effect of reverse diffusion steps on Urban100 ($\times4$).}
\label{tab:steps_ablation}
\footnotesize
\setlength{\tabcolsep}{5pt}
\renewcommand{\arraystretch}{0.95}
\begin{tabular}{cccc}
\toprule
Reverse steps & PSNR$\uparrow$ & SSIM$\uparrow$ & LPIPS$\downarrow$ \\
\midrule
50  & 28.11 & 0.8459 & 0.1698 \\
75  & 28.36 & 0.8487 & 0.1672 \\
100 & \best{28.53} & \best{0.8502} & \best{0.1647} \\
\bottomrule
\end{tabular}
\end{table}

\FloatBarrier

\section{Qualitative and Quantitative Results}
\label{sec:supp_qual}

This section provides additional qualitative comparisons on challenging urban scenes containing repeated windows, thin structural boundaries, facade patterns, and nighttime illumination. Such regions are particularly informative for assessing whether a super-resolution method can recover high-frequency detail while preserving geometric consistency.

\begin{figure*}[t]
    \centering
    \includegraphics[width=\textwidth]{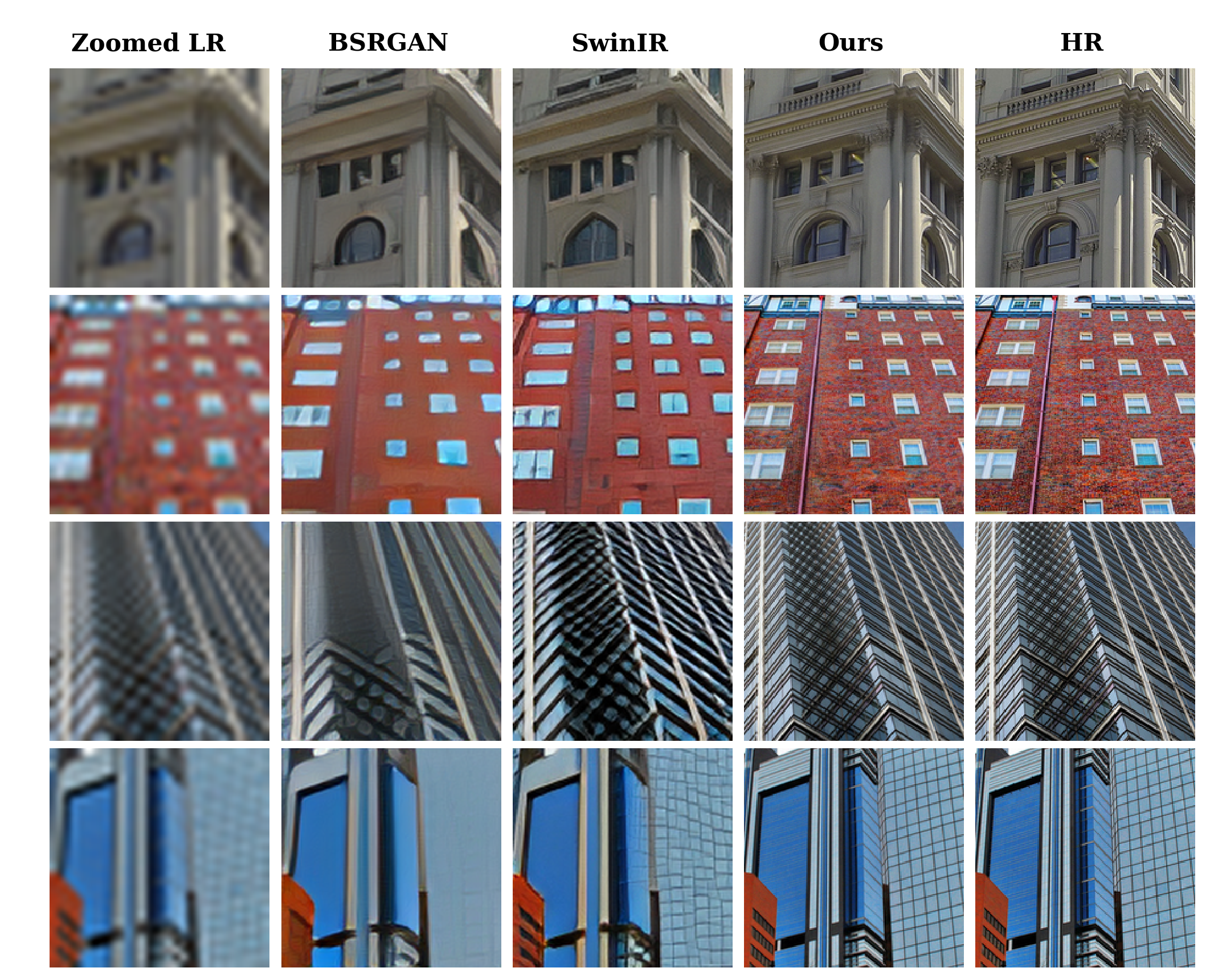}
    \caption{Qualitative comparison on Urban100 under the $\times8$ super-resolution setting.}
    \label{fig1}
\end{figure*}

Figure~\ref{fig1} presents four representative Urban100 examples under the $\times8$ setting, comparing several baseline methods (BSRGAN~\cite{BSRGAN}, SwinIR~\cite{SwinIR}), our BATDiff reconstruction, and the HR reference. The selected crops highlight structures that are typically difficult to reconstruct faithfully, including narrow window frames, repetitive brick layouts, dense line patterns, and sharp building contours.

Across these examples, BATDiff produces reconstructions that are visually closer to the HR reference, especially in regions with strong structural regularity. In the first example, the arch boundary and surrounding facade details are better restored, with sharper edge transitions and more coherent local contrast. In the second example, the repeated brick and window patterns are reconstructed more faithfully, whereas weaker baselines exhibit noticeable blurring or texture distortion. In the third and fourth examples, BATDiff preserves the continuity of fine line structures and building boundaries more effectively, reducing the structural breakup and over-smoothed appearance visible in lower-performing methods.

\subsection{Failure Cases}

\begin{figure*}[t]
    \centering
    \includegraphics[width=\textwidth]{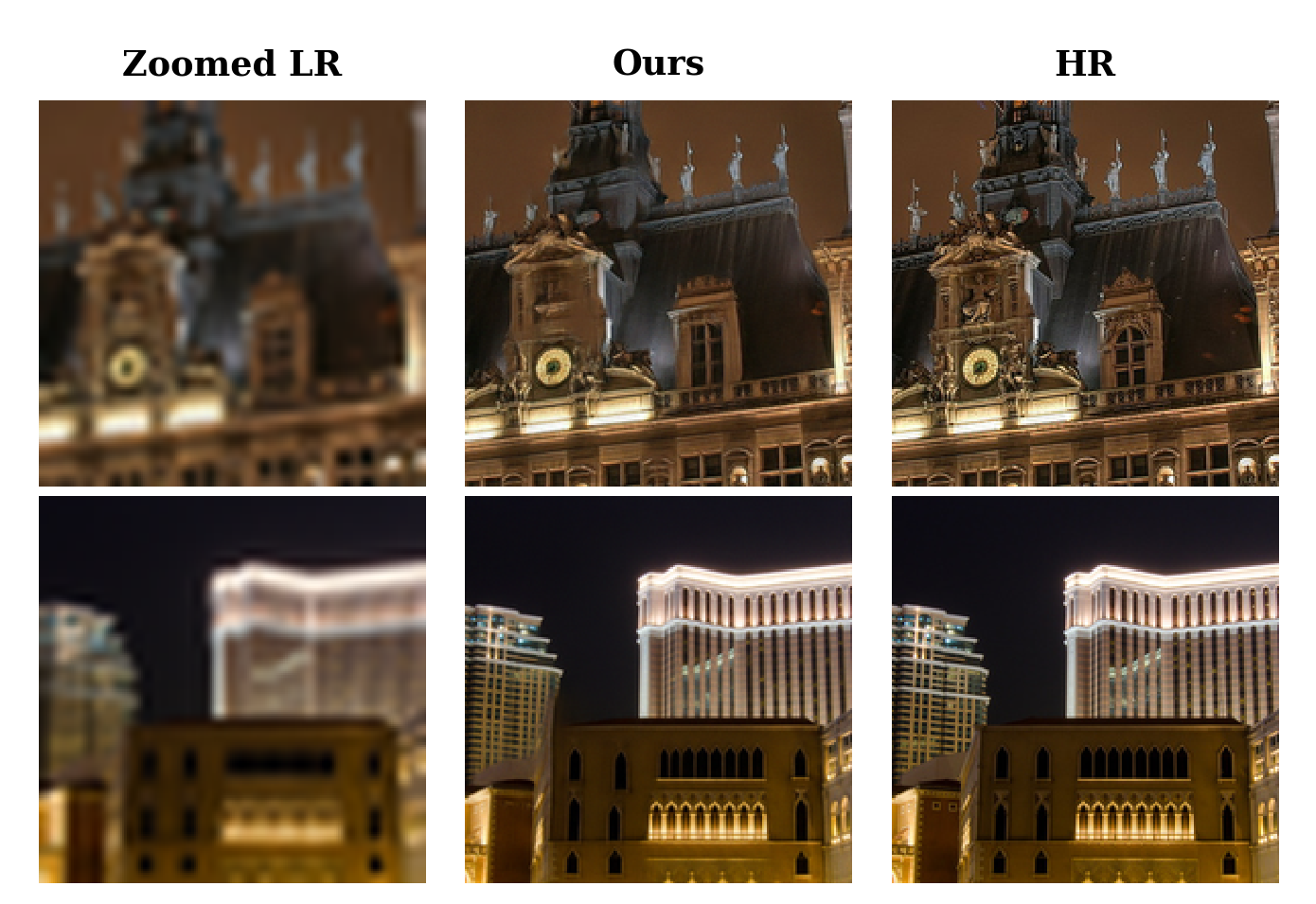}
    \caption{Representative failure cases under the $\times8$ setting. From left to right: Zoomed LR, ours, and HR reference. When the LR input contains extremely limited high-frequency information, the proposed method may produce slightly over-smoothed reconstructions or incomplete recovery of very fine architectural details.}
    \label{fig:supp_failure}
\end{figure*}

Although BATDiff generally improves structural reconstruction and texture fidelity, certain challenging cases remain difficult to recover reliably. Figure~\ref{fig:supp_failure} illustrates two representative examples under the $\times8$ setting.

In these scenes, the LR inputs contain extremely limited high-frequency information, particularly in regions with large smooth surfaces or weak internal repetition. As a result, the model cannot fully reconstruct the fine structural details present in the HR reference. In the first example, the thin architectural elements and decorative roof structures are only partially recovered, leading to slightly blurred boundaries compared with the ground truth. In the second example, strong illumination and smooth building surfaces make it difficult to infer missing high-frequency patterns, resulting in a mild loss of structural sharpness.

Nevertheless, even in these challenging scenarios, the proposed method preserves the global geometry and major structural components of the scene without introducing strong hallucinated artifacts.

\subsection{Evaluation at Multiple Scale Factors}

To further assess the flexibility of BATDiff under different super-resolution factors, we evaluated the proposed model in the $\times6$ setting. Table~\ref{tab:main_sr_x6} reports PSNR comparisons on Set5~\cite{Set5}, Set14~\cite{Set14}, and Urban100~\cite{Urban100} against representative arbitrary-scale SR approaches, including MetaSR~\cite{MetaSR}, LIIF~\cite{LIIF}, LTE~\cite{LTE}, CLIT~\cite{CLIT}, CiaoSR~\cite{CiaoSR}, and HIIF~\cite{HIIF}. To study the performance of BATDiff under non-integer scaling ratios, we also consider magnification factors of $\times3.15$ and $\times5.20$. The quantitative results on Set5, Set14, and Urban100 are summarized in Table~\ref{tab:continuous_scales}.

These results suggest that the multiscale diffusion formulation of BATDiff provides stable reconstruction behaviour even under intermediate scale factors. The cross-scale conditioning mechanism further helps maintain structural consistency when reconstructing high-frequency details at larger magnification levels.

\begin{table}[t]
\centering
\caption{Quantitative comparison for $\times6$ super-resolution (PSNR).}
\label{tab:main_sr_x6}
\footnotesize
\setlength{\tabcolsep}{2.4pt}
\renewcommand{\arraystretch}{0.88}
\begin{tabular}{l|ccccccc}
\toprule
Dataset & MetaSR~\cite{MetaSR} & LIIF~\cite{LIIF} & LTE~\cite{LTE} & CLIT~\cite{CLIT} & CiaoSR~\cite{CiaoSR} & HIIF~\cite{HIIF} & Ours$\times6$ \\
\midrule
Set5 & 28.74 & 28.93 & 28.93 & 28.95 & 29.10 & \second{29.17} & \best{29.86} \\
Set14 & 26.51 & 26.64 & 26.71 & 26.83 & 26.79 & \second{26.86} & \best{28.54} \\
Urban100 & 23.99 & 24.20 & 24.28 & 24.44 & 24.58 & \second{24.59} & \best{24.66} \\
\bottomrule
\end{tabular}
\end{table}

\begin{table}[t]
\centering
\caption{Performance of BATDiff for continuous super-resolution scales (PSNR).}
\label{tab:continuous_scales}
\footnotesize
\setlength{\tabcolsep}{4pt}
\renewcommand{\arraystretch}{0.9}
\begin{tabular}{c|lc}
\toprule
Scale & Dataset & PSNR$\uparrow$ \\
\midrule
\multirow{3}{*}{$\times3.15$}
 & Set5     & 33.26 \\
 & Set14    & 30.41 \\
 & Urban100 & 28.79 \\
\midrule
\multirow{3}{*}{$\times5.20$}
 & Set5     & 30.35 \\
 & Set14    & 28.96 \\
 & Urban100 & 25.11 \\
\bottomrule
\end{tabular}
\end{table}

\clearpage  

\bibliographystyle{splncs04}
\bibliography{main}
\end{document}